\documentclass[runningheads]{llncs}

\usepackage{hyperref}

\usepackage{amsmath,amsfonts,amssymb}

\usepackage{orcidlink}
\usepackage{graphicx}
\usepackage{caption}
\usepackage{subfig}
\usepackage{multicol}

\usepackage{algorithm}
\usepackage{algpseudocode}

\usepackage{tikz}
\usetikzlibrary{positioning}
\usetikzlibrary{arrows}
\usetikzlibrary{tikzmark}
\usetikzlibrary{backgrounds}
\usetikzlibrary{arrows.meta}
\usetikzlibrary{shadows} \usetikzlibrary{fadings}
\usetikzlibrary{shapes.arrows}
\usetikzlibrary{patterns}

\usetikzlibrary{3d}
\usepackage{pgfplots}
\pgfplotsset{compat=newest}

\usepackage{colortbl}

\usepackage{array}
\usepackage{diagbox}
\usepackage{multirow}
\usepackage{makecell}
\usepackage[export]{adjustbox}
\usepackage{subcaption}

\DeclareMathOperator*{\argmax}{arg\,max}
\DeclareMathOperator*{\argmin}{arg\,min}

\begin{document}
\title{Embedding Non-Distortive Cancelable Face Template Generation}

\titlerunning{Embedding Non-Distortive Template Generation}

\author{Dmytro Zakharov\inst{1}\orcidlink{0000-0001-9519-2444} \and
Oleksandr Kuznetsov\inst{1,2}\orcidlink{0000-0003-2331-6326} \and
Emanuele Frontoni\inst{2}\orcidlink{0000-0002-8893-9244}\and
Natalia Kryvinska \inst{3}\orcidlink{0000-0003-3678-9229}}
\authorrunning{D. Zakharov et al.}
%
\institute{
V.N. Karazin Kharkiv National University, 4 Svobody Sq., Kharkiv, 61022, Ukraine
\email{kuznetsov@karazin.ua, zamdmytro@gmail.com}\and
University of Macerata, Via Crescimbeni, 30/32, Macerata, 62100, Italy \email{emanuele.frontoni@unimc.it}\and
Comenius University Bratislava, Odbojárov 10, 820-05 Bratislava, Slovakia
\email{natalia.kryvinska@uniba.sk}}

\maketitle              

\begin{figure*}
\captionsetup{belowskip=-20pt}

\begin{center}
\begin{tabular}{cccccccc}
Real & Template & Real & Template & Real & Template & Real & Template \\ 
\includegraphics[width=1cm]{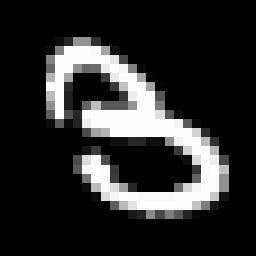} & \includegraphics[width=1cm]{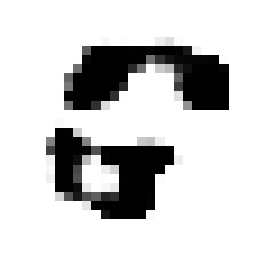} & \includegraphics[width=1cm]{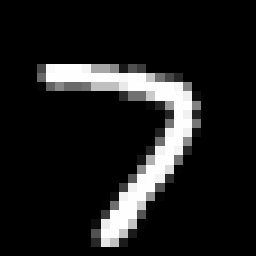}& \includegraphics[width=1cm]{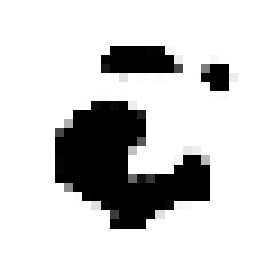} & \includegraphics[width=1cm]{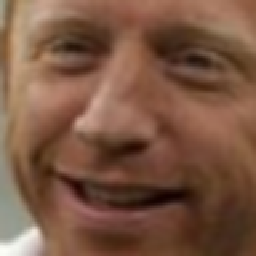} & \includegraphics[width=1cm]{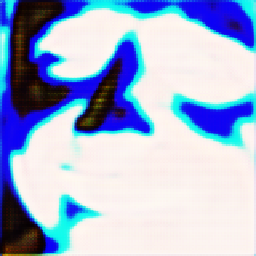} & \includegraphics[width=1cm]{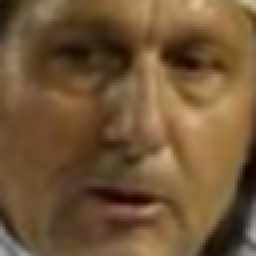} & \includegraphics[width=1cm]{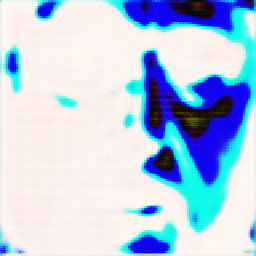} \\
\includegraphics[width=1cm]{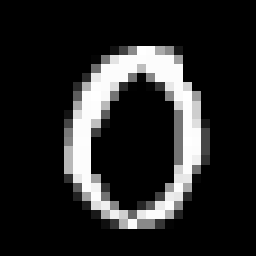} & \includegraphics[width=1cm]{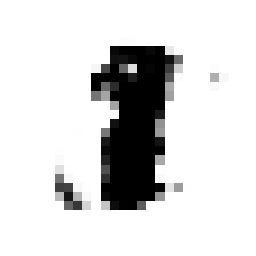} & \includegraphics[width=1cm]{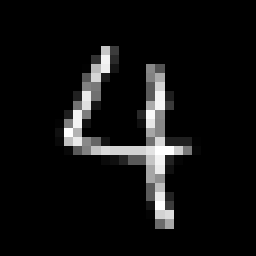}& \includegraphics[width=1cm]{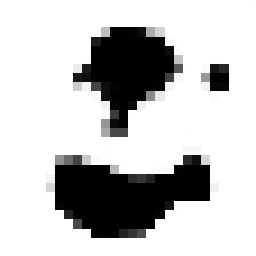} & \includegraphics[width=1cm]{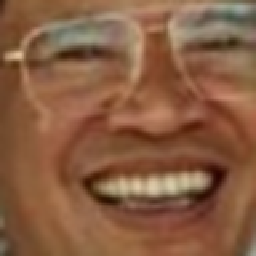} & \includegraphics[width=1cm]{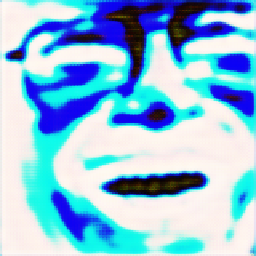} & \includegraphics[width=1cm]{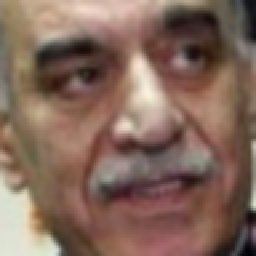} & \includegraphics[width=1cm]{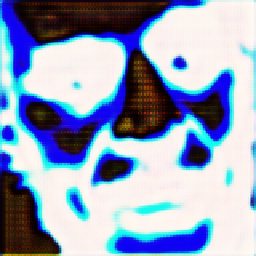} \\
\end{tabular}
\label{table:examples}
\end{center}

\caption{An example of using our proposed image distortion technique on images from \textit{MNIST} \cite{mnist} and \textit{LFW} \cite{lfw} datasets. While authentic and generated images significantly differ, the feature vectors of both images in pairs are relatively close.}
\end{figure*}

\begin{abstract}

Biometric authentication systems are crucial for security, but developing them involves various complexities, including privacy, security, and achieving high accuracy without directly storing pure biometric data in storage. We introduce an innovative image distortion technique that makes facial images unrecognizable to the eye but still identifiable by any custom embedding neural network model. Using the proposed approach, we test the reliability of biometric recognition networks by determining the maximum image distortion that does not change the predicted identity. Through experiments on \textit{MNIST} and \textit{LFW} datasets, we assess its effectiveness and compare it based on the traditional comparison metrics. 

\end{abstract}

\section{Introduction}

Biometric authentication systems have become a crucial topic in the development of modern cybersecurity solutions, offering a unique and personal way of identifying individuals \cite{authentication_techniques}. The integration of Artificial Intelligence (AI) methods has significantly advanced the field, enabling systems to accurately and efficiently process complex biometric data such as fingerprints, facial features, and iris patterns \cite{Stamp_Aaron_Visaggio_Mercaldo_Di_Troia_2022}. This progress has not only enhanced the precision of biometric systems but also their adaptability and scalability across diverse applications \cite{Puech_2022}.

However, the traditional biometric authentication systems face a critical vulnerability: the risk of biometric data compromise \cite{vulnerabilities_biometric}. Since biometric attributes are inherently unique and immutable, their theft or unauthorized access poses irreversible security risks. In conventional systems, biometric data must be stored, often in a digital format that, if intercepted, could lead to permanent identity theft, as these attributes cannot be reissued or altered like passwords.

In response to this challenge, the concept of \textit{cancelable biometrics} emerged as a transformative solution. This approach involves intentionally distorting biometric data before storage, ensuring that the stored information is a transformable representation rather than the original biometric signature. This method significantly mitigates the risk of compromising personal biometric data by making the stored information unusable outside the context of the specific authentication system for which it was designed.

Despite its advantages, cancelable biometrics introduces new complexities \cite{Yang_Wang_Shahzad_Zhou_2021}. Each device or system must implement its distortion mechanism, as authentication involves comparing distorted versions of biometric data. This requirement can complicate the deployment and interoperability of biometric systems across different platforms and devices.

Our work introduces a novel concept in this domain: \textit{Non-Distortive Cancelable Biometrics}. This innovative approach retains the benefits of cancelable biometrics but allows for the comparison of distorted biometric data against original human biometric features. We achieve this by designing AI models capable of recognizing the equivalence between original and intentionally altered biometric images. To the AI, these distorted and original images are perceptually close, while to any external observer without knowledge of the internal metrics used by the AI, they appear entirely different. This paradigm shift enables us to ``retrain'' AI perception, similarly to how certain animals perceive the world in a spectrum invisible to humans, thereby opening new avenues in cybersecurity applications.


\section{Background}

Currently, in our view, three primary image security directions are relevant to our study \cite{encoding_image_3}:
\begin{enumerate}
    \item \textbf{Steganography:} Concealing specified information inside the image with the further ability to retrieve this information \cite{steganography_overview,hiding_image}. This problem was successfully achieved using deep learning models \cite{steganography_overview,zhang2019steganogan}.
    \item \textbf{Cancelable Biometrics:} Image-to-image function which maps images to the unrecognizable format such that finding the inverse is computationally infeasible. When two photos of the same person get transformed, they get similar representations in the ``concealed'' space.
    \item \textbf{Image Encryption:} Similarly to ciphers, encoding and decoding functions, transforming the image from the plain version to the unrecognizable format and vice versa, respectively \cite{encoding_image_3,encoding_image_1,encoding_image_2}. Note that in this case, even small permutations in input produce drastically different outputs.
\end{enumerate}

While steganography seems promising, image encryption has no principal difference from the biometric system implementation perspective since we still need to conduct encoding/decoding operations. Therefore, we compare our method with the latter two options.

Now, suppose we are building the biometric authorization service. When the user wants to register, he passes the biometrics data (in our particular case, the face photo $p$), and using the generator $G$, we save the template $t=G(p)$ in the database. After that, some user passes a photo $p'$, and the system wants to verify that the user exists in the database $\mathcal{T}$ consisting of generated templates. Specifically, we need an algorithm to check whether the template $t \in \mathcal{T}$ belongs to the same person as $p'$. Here comes the main difference between the methods considered in Algorithm \ref{alg:auth}.

\begin{algorithm}
    \caption{Simplest approaches to authorization flows}
    \label{alg:auth}
    \begin{multicols}{3}
    \begin{algorithmic}
        \State \underline{\textbf{Cancel. biometrics}}
        \State Scan photo $p$
        \State $t_p \gets G(p)$
        \For{$t \in \mathcal{T}$}
            \State \textit{Accept} if $d(t_p,t) < \theta$
        \EndFor
        \textit{Reject}
    \end{algorithmic}
    \columnbreak
    \begin{algorithmic}
        \State \underline{\textbf{Encryption}}
        \State Scan photo $p$
        \For{$t \in \mathcal{T}$}
            \State $p_t \gets G^{-1}(t)$
            \State \textit{Accept} if $d(p,p_t) < \theta$
        \EndFor
        \State \textit{Reject}
    \end{algorithmic}
    \columnbreak
    \begin{algorithmic}
        \State \underline{\textbf{Our proposal}}
        \State Scan photo $p$
        \For{$t \in \mathcal{T}$}
            \State \textit{Accept} if $d^*(p,t) < \theta$
        \EndFor
        \State \textit{Reject}
    \end{algorithmic}
  \end{multicols}
\end{algorithm}

Consider the first two cases from Algorithm \ref{alg:auth}. For both cases, we compare two plain face images using metrics $d$, usually used in most face recognition systems. Note that in both cases, however, before calculating $d(\cdot,\cdot)$, we need to transform images in some way (either by $G$ or $G^{-1}$). 

The novelty of our approach, depicted as the third case in Algorithm \ref{alg:auth}, consists in the ability to directly compare $p$ and $t$ using the secret comparison metrics $d^*$. In other words, instead of conducting costly generation operations and applying traditional comparison methods, we use $d^*$ directly while maintaining high security by making $d(p,t)$ very large.

\section{Methodology}
\subsection{Embedding Model} \label{section:embedding}

Denote by $\mathcal{P}$ a set of photos (for example, for an RGB image we have $\mathcal{P} = \mathbb{R}^{W \times H \times n_C}$). In many applications, including face recognition, it is impractical to conduct operations directly with elements from $\mathcal{P}$: even $200 \times 200$ RGB image contains $120k$ individual elements. For that purpose, we introduce the term \textit{embedding model}, which can represent an element from $\mathcal{P}$ using a relatively low number of numbers.

\begin{definition}
    \textbf{Embedding Model}  is a function $E: \mathcal{P} \to \mathbb{R}^N$, which maps an image to a low-dimensional representation in $\mathbb{R}^N$ (typically for $N \lesssim 1024$), called \textbf{embedding} or sometimes \textbf{feature vector}.
\end{definition}

Besides, instead of having $\mathbb{R}^N$ as an embedding space, we decided to use the unit $N$-dimensional hypersphere $S^{N-1} \triangleq \{e \in \mathbb{R}^N: \|e\|_{2} = 1\}$ similarly to \textit{FaceNet} \cite{facenet} to boost performance (see \cite{normface} for details). Naturally, if we take two photos $p_1,p_2 \in \mathcal{P}$ of the same person, we should expect $E(p_1) \approx E(p_2)$. If, in turn, $p_2$ belongs to the different person than $p_1$, $E(p_1)$ and $E(p_2)$ should significantly differ. To concretize the metric of dissimilarity, we define the \textit{embedding distance}:
\begin{equation}\label{eq:emb_distance}
d_E(p_1,p_2) \triangleq \|E(p_1)-E(p_2)\|_2^2,
\end{equation}
where by $\|x\|_2^2 \triangleq \sum_{i=1}^d x_i^2$ we denote the $\ell_2$ norm. 

Now, we form the dataset $\mathbb{T} = \{(p^{\langle i \rangle},p^{\langle i \rangle}_+,p^{\langle i \rangle}_-)\}_{i=0}^{n_T} \subset \mathcal{P}^3$ where $p^{\langle i \rangle}$ and $p^{\langle i \rangle}_+$ belong to the same person (called \textit{anchor} and \textit{positive} images, respectively) while $p^{\langle i \rangle}_-$ to distinct people (called a \textit{negative} image). The idea of a triplet loss is to make embedding distance between anchor and negative larger than between anchor and positive. However, this condition on its own does not produce sufficiently good results, so we make an additional restriction: we want to have anchor-negative distance larger than an anchor-positive by a \textit{margin} $\gamma \in \mathbb{R}_{\geq 0}$. That being said, ideally, for any $(p,p_+,p_i) \in \mathbb{T}$ we want:
\begin{equation}
\gamma + d_E(p,p_+) < d_E(p,p_-).
\end{equation}
Formally speaking, suppose that we sample triplets $T=(p,p_+,p_-)$ from a true data distribution $\rho_{\text{data}}$. Our goal is to find such embedding model $E^*$ that maximizes the probability of the previously mentioned relationship:
\begin{equation}
    E^* = \argmax_{E}\mathbb{P}_{\langle p,p_+,p_- \rangle \sim \rho_{\text{data}}}\left[\gamma+d_{E}(p,p_+) < d_{E}(p,p_-) \right]
\end{equation}
Since this task is complicated to be directly solved, instead one considers the \textit{triplet loss function} to be minimized:
\begin{equation}\label{eq:triplet_loss}
\mathcal{L}_{\text{triplet}}(E)\triangleq \mathbb{E}_{\langle p,p_+,p_- \rangle \sim \rho_{\text{data}}}\left[\text{ReLU}\left(d_{E}(p,p_+) - d_{E}(p,p_-) + \gamma\right)\right],
\end{equation}
and thus we choose $E^*=\argmin_E \mathcal{L}_{\text{triplet}}(E)$.

\subsection{Generator Model}

\begin{definition}
\textbf{Distortion generator} is a function $G: \mathcal{P} \to \mathcal{P}$, which generates a distorted photo from a given one. This generator must meet the following two criteria:
\begin{enumerate}
    \item Difference between photos $G(p)$ and $p$ is as large as possible. We call the metrics for such difference $d_{\text{img}}: \mathcal{P} \times \mathcal{P} \to \mathbb{R}_{\geq 0}$.
    \item Difference between embeddings $E(G(p))$ and $E(p)$ is as small as possible. We call the metrics for this difference $d_{\text{emb}}: \mathbb{R}^N \times \mathbb{R}^N \to \mathbb{R}_{\geq 0}$.
\end{enumerate}
\end{definition}
Again, define the true distribution by $\rho_{\text{data}}$. Informally, the above definition can be also formulated as:
\begin{gather}\label{eq:informal_optimization}
\max_{G} \mathbb{E}_{p \sim \rho_{\text{data}}}\left[d_{\text{img}}(G(p), p)\right] \; \text{while} \; \min_{G} \mathbb{E}_{p \sim \rho_{\text{data}}}\left[ d_{\text{emb}}(E(G(p)), E(p))\right].
\end{gather}

Based on the formulation above, one might instantly think of GAN \cite{gan} since we have a pair of models $\langle E,G\rangle$ and a condition with $\min$ and $\max$. However, besides the fact than we assume $E$ to be fixed, there is a drastic difference -- here, $G$ does not want to fool $E$, but conversely take the random variable $G(p)$ as far as possible from $\rho_{\text{data}}$. However, the idea of training both $\langle E,G \rangle$ via a single setup is a great topic for further studies.

\subsection{Loss Function}\label{section:loss}

To represent the optimization problem above, we define the following loss function:
\begin{equation}\label{eq:loss_trainer}
\mathcal{L}(E,G) \triangleq (1-\beta) \cdot \mathcal{L}_{\text{img}}(G) + \beta\cdot\mathcal{L}_{\text{emb}}(E,G),
\end{equation}
where $\beta \in [0,1]$ is a positive hyperparameter, regulating the importance of $\mathcal{L}_{\text{emb}}$ in contrast to $\mathcal{L}_{\text{img}}$. 

We define the two loss components as follows:
\begin{gather}
\mathcal{L}_{\text{img}}(G) \triangleq -\mathbb{E}_{p \sim \rho_{\text{data}}}\left[d_{\text{img}}(G(p), p)\right],\\ \mathcal{L}_{\text{emb}}(E,G) \triangleq \mathbb{E}_{p \sim \rho_{\text{data}}}\left[\text{ReLU}(d_{\text{emb}}(E(G(p)), E(p))-\alpha)\right].
\end{gather}

Note that $\mathcal{L}_{\text{img}}$ is always negative since we want to \textit{maximize} the difference between images. Also, we decide to use $\text{ReLU}(d_{\text{emb}}(\cdot)-\alpha)$ for $\mathcal{L}_{\text{emb}}$ instead of $d_{\text{emb}}(\cdot)$ since otherwise neural network might focus primarily on reducing the distance between embeddings. However, if we use the ReLU function, we do not punish the neural network for an embedding difference unless it exceeds $\alpha$. In this sense, $\alpha$ also serves as a parameter that regulates how well we want our generator to fit embeddings: the larger $\alpha$ is, the more distinct images are according to metrics $d_{\text{img}}$, but less similar according to $d_{\text{emb}}$.

Let us now choose the concrete expressions for distances. We use $d_{E}$ from \autoref{eq:emb_distance} for the embedding difference:
\begin{equation}
d_{\text{emb}}(X;G,E) \triangleq d_{E}(G(p), p) = \|E(G(p)) - E(p)\|_2^2.
\end{equation}
Choosing $d_{\text{img}}$ is trickier. Below, we discuss several choices:
\begin{itemize}
    \item \textbf{Mean-squared error (MSE):} A frequent choice in many applications \cite{shadow_removal_l2,style_transfer}, which is defined as the mean value of squared differences pixelwise. 
    \item \textbf{Hamming Distance}. Similarly to MSE, the Hamming distance is used in many applications (see \cite{pix2pix,shadow_removal_l1_1,shadow_removal_l1_2}), yet in contrast to MSE, it encourages less blurring. It is defined as the mean value of absolute differences pixelwise. 
    \item \textbf{DSSIM.} To account for structural difference instead of an elementwise one, based on suggestion by \cite{image_loss}, we define the dissimilarity as $\frac{1-\text{SSIM}(X,Y)}{2}$, where 
    \[
    \text{SSIM}(X,Y) \triangleq \frac{(2\mu_X\mu_Y + \epsilon_1)(2\sigma_{XY}+\epsilon_2)}{(\mu_X^2+\mu_Y^2 + \epsilon_1)(\sigma_X^2+\sigma_Y^2+\epsilon_2)}.
    \]
    Here, $\mu_X,\mu_Y$ are pixel sample means, $\sigma_X^2,\sigma_Y^2$ -- variances, $\sigma_{XY}$ is a covariance, and finally $\epsilon_1,\epsilon_2$ are constants to stabilize the division.
    \item \textbf{Sobel Distance.} To focus increasing the difference on edges, for photo $p$ and prediction $\hat{p}$, we define the Sobel distance as the Hamming distance between $\mathcal{S} \odot p$ and $\mathcal{S} \odot \hat{p}$, where $\mathcal{S}$ is the mask, which is calculated by applying the Sobel filter on the true image $p$. Since $\mathcal{S}$ does not depend on the prediction $\hat{p}$, the value of mask might be generated prior to the training session.
    \item \textbf{Combined Distance.} We define it as the linear combination of aforementioned distances. We achieved the best results using the Hamming distance equipped with the Sobel distance. That being said, the distance is defined as $d_{\text{img}}(\hat{p},p) \triangleq \gamma \cdot d_{H}(\hat{p},p) + (1-\gamma) \cdot d_{S}(\hat{p},p)$, where $d_H$ and $d_S$ are Hamming and Sobel distances, respectively. In our experiments, we use $\gamma = 0.5$. Note that other combinations might be applied.
\end{itemize}

\subsection{Trainer Network Architecture}

When we finally defined the loss $\mathcal{L}(E,G)$, we need to train our generator to minimize this expected loss, that is $G^* = \argmin_{G}\mathcal{L}(E,G)$.

\begin{figure}
    \centering
    \begin{tikzpicture}

    \node[very thick, draw=black, fill=gray, fill opacity=0.2,minimum width=1.5cm, minimum height=4.5cm,dashed,on background layer](targetbox) at (8.5,-1.25) {};
    \node(targetzonelabel)[right=0.6cm of targetbox,rotate=-90,anchor=north]{Target image + embedding};
    
    \node[anchor=center,inner sep=0](realimg) at (0,0) {\includegraphics[width=0.1\textwidth]{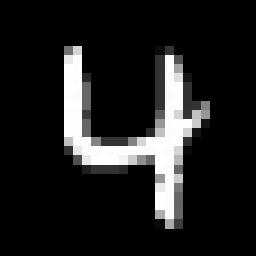}};
    \node[anchor=center,inner sep=0](generatedimg) at (5,0) {\includegraphics[width=0.1\textwidth]{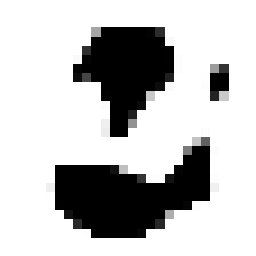}};
    \node[anchor=center,inner sep=0](targetimg) at (8.5,0) {\includegraphics[width=0.1\textwidth]{trainer_real.png}};

    \node[draw,shape=rectangle,color=purple!60,fill=purple!2,drop shadow,minimum size=1cm,very thick](gen) at (2.5,0) {\huge $\boldsymbol{G}$};
    \node[draw,shape=rectangle,color=blue!60,fill=blue!2,drop shadow,minimum size=1cm,very thick](emb) at (5,-2.5) {\huge $\boldsymbol{E}$};

    \node[draw,shape=rectangle,color=black!60,fill=black!2,drop shadow,minimum width=0.5cm, minimum height=1.5cm, very thick](embvecgen) at (6.5,-2.5) {};
    \node[draw,shape=circle,color=black!60,fill=black!2,minimum width=0.3cm, minimum height=0.3cm, very thick](embvecgen1) at (6.5,-2.5) {};
    \node[draw,shape=circle,color=black!60,fill=black!2,minimum width=0.3cm, minimum height=0.3cm, very thick](embvecgen2) at (6.5,-2) {};
    \node[draw,shape=circle,color=black!60,fill=black!2,minimum width=0.3cm, minimum height=0.3cm, very thick](embvecgen3) at (6.5,-3) {};
    \node[draw,shape=rectangle,color=black!60,fill=black!2,drop shadow,minimum width=0.5cm, minimum height=1.5cm, very thick](embvecreal) at (8.5,-2.5) {};
    \node[draw,shape=circle,color=black!60,fill=black!2,minimum width=0.3cm, minimum height=0.3cm, very thick](embvecreal1) at (8.5,-2.5) {};
    \node[draw,shape=circle,color=black!60,fill=black!2,minimum width=0.3cm, minimum height=0.3cm, very thick](embvecreal2) at (8.5,-2) {};
    \node[draw,shape=circle,color=black!60,fill=black!2,minimum width=0.3cm, minimum height=0.3cm, very thick](embvecreal3) at (8.5,-3) {};

    \node(reallabel)[above=0.05 of realimg]{Input $p$};
    \node(generatedlabel)[above=0.05 of generatedimg]{Generated image $G(p)$};
    \node[align=center](generatedlabel)[below=0.05 of embvecgen]{Embedding\\$E(G(p))$};

    \node[align=center](genmodlabel)[below=0.05 of gen]{Generator\\model};
    \node[align=right](embmodlabel)[left=0.15 of emb]{Embedding\\model};

    \path[line width=0.5mm](realimg) edge [-{Stealth[length=3mm]}] node {} (gen);
    \path[line width=0.5mm](gen) edge [-{Stealth[length=3mm]}] node {} (generatedimg);
    \path[line width=0.5mm](generatedimg) edge [-{Stealth[length=3mm]}] node {} (emb);
    \path[line width=0.5mm](emb) edge [-{Stealth[length=3mm]}] node {} (embvecgen);
    \path[line width=0.5mm,above,color=purple!60](generatedimg) edge [{Stealth[length=3mm]}-{Stealth[length=3mm]}] node {$d_{\text{img}}$} (targetimg);
    \path[line width=0.5mm,above,color=blue!60](embvecgen) edge [{Stealth[length=3mm]}-{Stealth[length=3mm]}] node {$d_{\text{emb}}$} (embvecreal);

    \end{tikzpicture}
    \caption{Trainer Network architecture}
    \label{fig:trainer_network}
\end{figure}

To achieve this, inspired by \cite{inverting}, we create a helper network, which we call a \textit{Trainer Network}. Its architecture is depicted in \autoref{fig:trainer_network}.

For training, we form the dataset in which photo $p \in \mathcal{P}$ is an input while a pair of the same photo and its embedding $\langle p,E(p)\rangle \in \mathcal{P} \times \mathbb{R}^N$ is an output (additionally, one might predefine Sobel masks $\{\mathcal{S}_i\}$ at this step to avoid calculating them online).

The trainer network takes a photo $p$, generates an image $G(p)$, and then takes the embedding of this image $E(G(p))$. It then outputs both values and applies the loss from \autoref{eq:loss_trainer}.

\section{Implementation}

\subsection{Embedding model}
For the \textit{LFW} dataset, we employed the \textit{FaceNet} architecture with previously trained weights since it achieves excellent accuracy on \textit{LFW}: 98.87\% for fixed center cropping, and 99.63\% for the extra face alignment (see original paper \cite{facenet} for details). Note that any other custom embedding neural network might have been used, such as \textit{VGGFace} \cite{vggface}, for example.

\begin{figure}
    \centering
    \begin{tikzpicture}        
        \node[anchor=center,inner sep=0](img) at (0,0) {\includegraphics[width=0.125\textwidth]{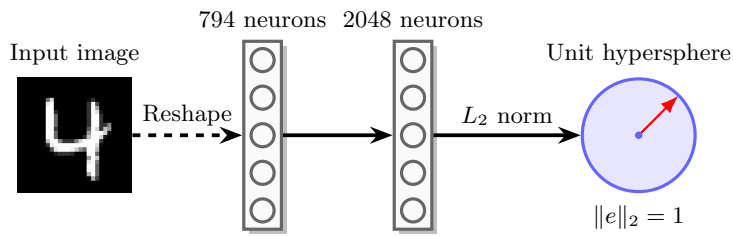}};

        \node[draw,shape=rectangle,color=black!60,fill=black!2,drop shadow,minimum width=0.5cm, minimum height=2.5cm, very thick](reshape) at (2.5,0) {};
        \node[draw,shape=circle,color=black!60,fill=black!2,minimum width=0.3cm, minimum height=0.3cm, very thick](reshape1) at (2.5,0) {};
        
        \node[draw,shape=circle,color=black!60,fill=black!2,minimum width=0.3cm, minimum height=0.3cm, very thick](reshape2) at (2.5,-1) {};
        \node[draw,shape=circle,color=black!60,fill=black!2,minimum width=0.3cm, minimum height=0.3cm, very thick](reshape3) at (2.5,-0.5) {};
        \node[draw,shape=circle,color=black!60,fill=black!2,minimum width=0.3cm, minimum height=0.3cm, very thick](reshape4) at (2.5,0.5) {};
        \node[draw,shape=circle,color=black!60,fill=black!2,minimum width=0.3cm, minimum height=0.3cm, very thick](reshape5) at (2.5,1) {};

        \node[draw,shape=rectangle,color=black!60,fill=black!2,drop shadow,minimum width=0.5cm, minimum height=2.5cm, very thick](hidden) at (4.5,0) {};
        \node[draw,shape=circle,color=black!60,fill=black!2,minimum width=0.3cm, minimum height=0.3cm, very thick](hidden1) at (4.5,0) {};
        
        \node[draw,shape=circle,color=black!60,fill=black!2,minimum width=0.3cm, minimum height=0.3cm, very thick](hidden2) at (4.5,-1) {};
        \node[draw,shape=circle,color=black!60,fill=black!2,minimum width=0.3cm, minimum height=0.3cm, very thick](hidden3) at (4.5,-0.5) {};
        \node[draw,shape=circle,color=black!60,fill=black!2,minimum width=0.3cm, minimum height=0.3cm, very thick](hidden4) at (4.5,0.5) {};
        \node[draw,shape=circle,color=black!60,fill=black!2,minimum width=0.3cm, minimum height=0.3cm, very thick](hidden5) at (4.5,1) {};

        \node[draw,shape=circle,color=blue!60,fill=blue!10,minimum width=1.5cm, minimum height=1.5cm, very thick](unitcircle) at (7.5,0) {};
        \node (A) at (7.4, -0.1) {};
        \node (B) at (7.5+1.41421/2.2, 1.41421/2.2) {};
        \draw[-Latex,line width=0.3mm,color=red](A) edge (B);
        \node[circle,fill,inner sep=1pt,color=blue!60] at (7.5,0){};

        \path[line width=0.5mm,dashed](img) edge [-{Stealth[length=3mm]},above] node {Reshape} (reshape);
        \path[line width=0.5mm](reshape) edge [-{Stealth[length=3mm]},above] node {} (hidden);
        \path[line width=0.5mm](hidden) edge [-{Stealth[length=3mm]},above] node {$L_2$ norm} (unitcircle);

        \node(imglabel)[above=0.05 of img]{Input image};
        \node(reshapelabel)[above=0.05 of reshape]{794 neurons};
        \node(hiddenlabel)[above=0.05 of hidden]{2048 neurons};
        \node(unitcirclelabelbelow)[below=0.05 of unitcircle]{$\|e\|_2 = 1$};
        \node(unitcirclelabelabove)[above=0.05 of unitcircle]{Unit hypersphere};
    \end{tikzpicture}
    \caption{Embedding model architecture.}
    \label{fig:embedding-arch}
\end{figure}

In turn, for the \textit{MNIST} we decided to build and train our own model, architecture of which is specified in \autoref{fig:embedding-arch}. We use the LeakyReLU activation function with $\alpha=0.01$. For the output layer, we use the $L_2$ normalization with no activation function beforehand. We choose $N=10$ to be our embeddings dimensionality and a margin $\mu=0.2$. We then train our model using Adam optimizer \cite{adam} with a learning rate of $10^{-4}$.

\subsection{Generator model}

For the generator model, we decided to employ the \textit{U-Net} architecture
\cite{unet}. Similarly to the embedding model from the previous section, we use \textit{He} \cite{he} weights initialization, LeakyReLU activation for all convolutional layers except for the last one, and the sigmoid function before the output to map pixel values to the interval $(0,1)$. We use batch size of $64$ with a learning rate $10^{-4}$. Other parameters depend on the dataset:
\begin{itemize}
    \item For the \textit{MNIST} dataset, we use a margin $\alpha=0.3$, $\beta=0.9$, and MSE as the loss function.
    \item For the \textit{LFW} dataset, we use a margin $\alpha=0.2$, $\beta=0.1$, and the combined distance.
\end{itemize}

\section{Results}

In this section, we present the experimental results obtained from evaluating the proposed non-distortive cancelable biometric authentication system based on two widely recognized datasets: \textit{MNIST} \cite{mnist} and \textit{LFW} \cite{lfw} (Labeled Faces in the Wild), allowing us to assess the effectiveness of our method against established benchmarks using conventional comparison metrics.

\subsection{Hamming Distance Analysis}

To quantitatively evaluate the alterations introduced by our system and their impact on image recognizability while retaining identity verification capabilities, we applied the Hamming distance metric ($d_H$) as a fundamental dissimilarity measure. We simulated three scenarios, each highlighting a unique aspect of the authentication system's performance:

\begin{itemize}
    \item \textit{Real-Real Comparison:} This setup involves contrasting real images from different classes to assess the baseline variability of the dataset without any introduced distortions. 
    \item \textit{Real-Gen Comparison:} In this scenario, we juxtapose real images with their distorted counterparts generated by our system, while ensuring both belong to the same class (i.e., the same individual's biometric data). This comparison is crucial for quantifying the degree of alteration our distortion technique imposes on the images, thereby evaluating its impact on the recognizability and integrity of biometric data. 
    \item \textit{Gen-Gen Comparison:} Here, we compare generated images across different classes to understand how our distortion technique affects the differentiation between distinct identities after the distortion process. High dissimilarity (higher Hamming distances) in this scenario would suggest that our system effectively maintains the separability of different biometric identities, even after the application of distortion techniques. This aspect is vital for ensuring the system's robustness against potential impersonation attacks or misidentification errors.
\end{itemize}

\begin{figure}
\pgfplotsset{scaled y ticks=false}
\begin{minipage}{.45\textwidth}
\centering
\begin{tikzpicture}[scale=0.65]
\begin{axis}[
ybar,
grid,
ylabel=Frequency,
xlabel=Hamming distance $d_H$,
axis x line=bottom,
axis y line=left
]
\addplot+[ybar interval,mark=no] plot coordinates {(0.045,15)(0.068,95)(0.090,1066)(0.112,5767)(0.134,13036)(0.156,19107)(0.178,18266)(0.200,13549)(0.222,8822)(0.244,5048)(0.266,2836)(0.288,1032)(0.311,572)(0.333,477)(0.355,190)(0.377,43)(0.399,13)(0.421,61)(0.443,0)(0.465,5)};
\addplot+[ybar interval,mark=no]
            plot coordinates {(0.484,60)(0.499,480)(0.513,1890)(0.528,3780)(0.542,5840)(0.557,7460)(0.572,6660)(0.586,5460)(0.601,4050)(0.615,2650)(0.630,1780)(0.645,1010)(0.659,690)(0.674,450)(0.688,330)(0.703,80)(0.718,80)(0.732,80)(0.747,20)(0.761,10)};
\addplot+[ybar interval,mark=no,opacity=0.5,green!100,fill=green!35] plot coordinates {(0.068,33)(0.106,419)(0.144,1193)(0.182,3305)(0.220,6220)(0.259,10849)(0.297,12468)(0.335,14069)(0.373,13547)(0.411,10274)(0.449,7181)(0.487,4322)(0.525,2708)(0.563,1604)(0.602,949)(0.640,484)(0.678,253)(0.716,89)(0.754,29)(0.792,4)};
\legend{Real-Real,Real-Gen,Gen-Gen};
\end{axis}
\end{tikzpicture} \\
\textbf{(a)} LFW Dataset
\end{minipage}
\begin{minipage}{.45\textwidth}
\centering
\begin{tikzpicture}[scale=0.65]
\begin{axis}[
ybar,
grid,
ylabel=Frequency,
xlabel=Hamming distance $d_H$,
axis x line=bottom,
axis y line=left
]
\addplot+[ybar interval,mark=no] plot coordinates {(0.569,5)(0.585,11)(0.600,48)(0.615,212)(0.630,784)(0.646,1283)(0.661,1970)(0.676,3216)(0.691,4652)(0.707,5604)(0.722,7685)(0.737,9745)(0.752,9506)(0.768,6826)(0.783,4125)(0.798,2494)(0.813,1236)(0.829,468)(0.844,120)(0.859,10)};
\addplot+[ybar interval,mark=no]
            plot coordinates {(0.484,60)(0.499,480)(0.513,1890)(0.528,3780)(0.542,5840)(0.557,7460)(0.572,6660)(0.586,5460)(0.601,4050)(0.615,2650)(0.630,1780)(0.645,1010)(0.659,690)(0.674,450)(0.688,330)(0.703,80)(0.718,80)(0.732,80)(0.747,20)(0.761,10)};
\addplot+[ybar interval,mark=no,opacity=0.5,green!100,fill=green!35] plot coordinates {(0.026,400)(0.042,1470)(0.059,3000)(0.075,4130)(0.092,4110)(0.108,4090)(0.124,3450)(0.141,2930)(0.157,1950)(0.174,1730)(0.190,990)(0.207,650)(0.223,490)(0.240,240)(0.256,130)(0.273,90)(0.289,80)(0.305,30)(0.322,30)(0.338,10)};
\legend{Real-Real,Real-Gen,Gen-Gen};
\end{axis}
\end{tikzpicture} \\
\textbf{(b)} MNIST Dataset
\end{minipage}
\caption{Histogram of Hamming distances for three cases: ``Real vs Real photos'', ``Real vs Generated photos'', and ``Generated vs Generated photos'' and two different datasets: \textit{(a)} \textit{LFW}, \textit{(b)} \textit{MNIST}.}
\label{fig:hamming}
\end{figure}
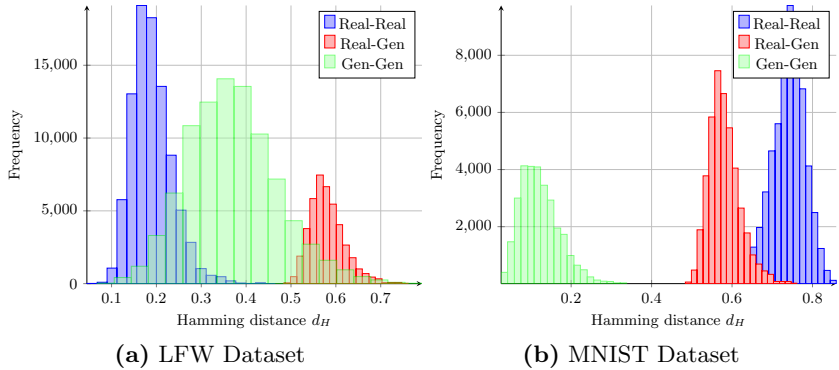

\autoref{fig:hamming} showcases the distribution of Hamming distances across the aforementioned scenarios for both LFW and MNIST datasets. Notably, the histograms reveal a distinct pattern of dissimilarity distributions, reflecting the efficacy of our distortion technique in preserving the identity information while significantly altering the visual appearance of the images.

\subsection{Test Recognition System}

To rigorously evaluate the performance of our proposed approach, we implemented a test recognition system. This system's primary aim is to determine the extent to which our image distortion method affects the accuracy of binary classification metrics typically used in authentication systems. By simulating a real-world authentication scenario, we aimed to understand the practical implications of our method on system performance.

That being said, first we implement the Algorithm \ref{alg:auth}: suppose the database $\mathcal{T}$ consists of distorted face images, and user with a photo $p$ tries to log in. We accept him if $\exists t \in \mathcal{T}: d_E(p,t) < \theta$.

To build the confusion matrix  we launch the Algorithm \ref{alg:rec_system}. This way, we get the true positive (TP), true negative (TN), false positive (FP), and false negative (FN) values. Since it is unclear which $\theta$ to pick, we try many different values in range $[0,4]$ and choose $\theta^*$ which yields the lowest \textit{average error rate} (see definition in \autoref{section:performance_metrics}).

\begin{algorithm}
\begin{algorithmic}
        \State \underline{\textbf{Preprocessing Stage}}
        \State 1. Get all distinct classes $\mathcal{C}$ (for MNIST dataset, for instance, we have $\mathcal{C} = \{0,\dots,9\}$).
        \State 2. Choose a random subset of classes $\mathcal{C}_{\text{valid}} \subset \mathcal{C}$ which would correspond to those users who have already ``registered'' in the system.
        \State 3. Sample $1$ person photo randomly (uniformly) for each class from $\mathcal{C}_{\text{valid}}$.
        \State 4. Apply pre-trained $G$ to each retrieved photo and put into the database $\mathcal{T}$.
        \State \underline{\textbf{Validation Stage}}
        \State 1. Sample $N=1000$ face images $\mathcal{P}_{\text{valid}}$ from $\mathcal{C}_{\text{valid}}$ except for those registered.
        \State 2. Sample $N$ face images $\mathcal{P}_{\text{invalid}}$ from $\mathcal{C} \setminus \mathcal{C}_{\text{valid}}$. 
        \State 3. $\text{TP} \gets \sum_{p \in \mathcal{P}_{\text{valid}}}\texttt{Auth}(p \mid \mathcal{T})$ \Comment{$\texttt{Auth}(p\mid \mathcal{T})$ is $1$ if authorization using photo $p$ is successful and $0$ otherwise.}
        \State 4. $\text{FN} \gets N - \text{TP}$
        \State 5. $\text{TN} \gets \sum_{p \in \mathcal{P}_{\text{invalid}}} (1-\texttt{Auth}(p \mid \mathcal{T}))$
        \State 6. $\text{FP} \gets N - \text{TN}$.  
        \State \Return $\langle \text{TP},\text{FN},\text{TN},\text{FP}\rangle$
    \end{algorithmic}
    \caption{Simulating the authentication system to estimate TP, TN, FP, and FN.}
    \label{alg:rec_system}
\end{algorithm}

\subsection{Performance Metrics Evaluation}\label{section:performance_metrics}

After getting TP, TN, FP, and FN values, we decided to employ the advanced performance metrics to understand the system's operational nuances, especially in the context of biometric authentication, where the balance between security and usability is essential. 

The metrics under consideration include \textit{Accuracy}, \textit{Specificity}, \textit{Negative Predictive Value}, \textit{False Positive Rate}, \textit{False Negative Rate}, and \textit{Average Error Rate}. Each metric offers certain system's performance insights:
\begin{itemize}
    \item \textbf{Accuracy (Acc)} measures the overall effectiveness of the system in classifying both positive and negative instances correctly.
    \item \textbf{Specificity (Sp)} measures the system's ability to correctly reject unauthorized users, a critical measure for ensuring security.
    \item \textbf{Negative Predictive Value (NPV)} reflects the likelihood that a negative classification truly means the absence of the condition, reinforcing trust in the system's decisions.
    \item \textbf{False Positive Rate (FPR)} and \textbf{False Negative Rate (FNR)} represent the probabilities of incorrectly accepting an unauthorized user and wrongly rejecting an authorized user, respectively, indicating the system's error tendencies.
    \item \textbf{Average Error Rate (AER)} provides a holistic view of the system's error propensity, combining the impact of both types of errors.
\end{itemize}

Given a confusion matrix with True Positives (TP), True Negatives (TN), False Positives (FP), and False Negatives (FN), the formulas for these metrics are:

\begin{align}
&\text{Acc} = \frac{\text{TP} + \text{TN}}{\text{TP} + \text{TN} + \text{FP} + \text{FN}}, \;& \text{Sp} = \frac{\text{TN}}{\text{TN} + \text{FP}},& \nonumber \\
&\text{NPV} = \frac{\text{TN}}{\text{TN} + \text{FN}}, \;& \text{FPR} = \frac{\text{FP}}{\text{FP} + \text{TN}},& \\ 
&\text{FNR} = \frac{\text{FN}}{\text{TP} + \text{FN}}, \;& \text{AER} = \frac{\text{FPR} + \text{FNR}}{2}.& \nonumber
\end{align}

The performance metrics for each dataset and condition are presented in Table \ref{table:comprehensive_metrics}, offering a view of the system's capabilities.

\begin{table}[htbp]
\centering
\caption{Performance Metrics across specified Datasets and Conditions}
\label{table:comprehensive_metrics}
\begin{tabular}{c|cc|cc}
\Xhline{3.5\arrayrulewidth}
\multirow{2}{*}{\textbf{Metric}} & \multicolumn{2}{c|}{\textbf{MNIST Dataset}} & \multicolumn{2}{c}{\textbf{LFW Dataset}} \\ \cline{2-5} 
                        & \textbf{No Distortion} & \textbf{  With Distortion} & \textbf{No Distortion} & \textbf{  With Distortion} \\ \hline
Accuracy          & 0.960             & 0.964          & 0.934             & 0.937          \\
Specificity        & 0.958              & 0.948           & 0.929             & 0.927          \\
NPV                     & 0.961             & 0.978          & 0.938            & 0.946         \\
FPR                     & 0.042              & 0.052           & 0.071             & 0.073          \\
FNR                     & 0.039              & 0.021           & 0.061             & 0.053          \\
AER                     & 0.041             & 0.037          & 0.066             & 0.063          \\ 
\Xhline{3.5\arrayrulewidth}
\end{tabular}
\end{table}

The results encapsulated in Table \ref{table:comprehensive_metrics} reveal several key insights into the performance of our authentication system. Notably, the application of our image distortion technique generally enhances the system's accuracy, as evidenced by the improved \textit{accuracy} values for both datasets when distortion is applied. This increment underscores the efficacy of the distortion in preserving essential biometric features while saving non-critical information, thus bolstering recognition accuracy.

\section{Discussion and Comparison of Results}
Our investigation into non-distortive cancelable biometric authentication systems has yielded promising results, demonstrating the viability of enhancing privacy without significantly compromising system accuracy and reliability. The observed improvements in accuracy and error rates, particularly with the application of distortion, underscore the effectiveness of our approach in maintaining the integrity of biometric verification processes while adding a layer of security through biometric data transformation.

Comparatively, existing research in the domain of cancelable biometrics has primarily focused on trade-offs between security enhancements and performance detriments. For instance, studies such as those by \cite{Kaur_Khanna_2020} or \cite{Yang_Wang_Kang_Johnstone_Bedari_2022} have shown that while cancelable biometrics can significantly improve privacy protection, they often do so at the cost of increased false acceptance and rejection rates. In contrast, our system demonstrates that it is possible to implement distortion techniques that not only preserve but in some cases enhance the system's performance metrics.

\section{Conclusion}
This study has presented a comprehensive evaluation of a novel non-distortive cancelable biometric authentication system, highlighting its potential to resolve the often conflicting demands of privacy protection and authentication performance. Through extensive experimentation and analysis, we have demonstrated that our proposed system not only meets but in some instances, exceeds the performance standards of traditional biometric systems, while introducing significant privacy and security enhancements.

Our findings indicate that the strategic application of image distortion techniques can indeed result in a more secure and efficient authentication process, paving the way for the next generation of biometric authentication solutions. As the field of biometrics continues to evolve, it is imperative that privacy and security considerations remain at the forefront of technological advancements. To this end, our research contributes valuable insights and methodologies that can inform future developments in cancelable biometrics, ensuring that user privacy and system integrity are preserved.

%
%

\bibliographystyle{splncs04}
\bibliography{refs}

\end{document}